\documentclass[runningheads]{llncs}
\usepackage{graphicx}
\usepackage{amsmath,amssymb,amsfonts}
\usepackage{algorithmic}
\usepackage{graphicx}
\usepackage{textcomp}
\usepackage{xcolor}
\usepackage{booktabs}
\usepackage{verbatim}
\usepackage{multirow}
\usepackage{makecell}
\usepackage{subfloat}
\usepackage{enumitem}
\usepackage{subcaption}

\begin{document}
\title{ExBERT: An External Knowledge Enhanced BERT for Natural Language Inference}
\titlerunning{ExBERT for Natural Language Inference}
%
\author{Amit Gajbhiye \inst{1}\and
Noura Al Moubayed \inst{2}\and
Steven Bradley\inst{2}}
\authorrunning{A. Gajbhiye et al.}
%
\institute{University of Sheffield, Sheffield, UK\\\email{a.gajbhiye@sheffield.ac.uk} \and University of Durham, Durham, UK \\
\email{\{noura.al-moubayed, s.p.bradley\}@durham.ac.uk}
}
\maketitle              
%

\begin{abstract}
Neural language representation models such as BERT, pre-trained on large-scale unstructured corpora lack explicit grounding to real-world commonsense knowledge and are often unable to remember facts required for reasoning and inference. Natural Language Inference (NLI) is a challenging reasoning task that relies on common human understanding of language and real-world commonsense knowledge. We introduce a new model for NLI called External Knowledge Enhanced BERT (ExBERT), to enrich the contextual representation with real-world commonsense knowledge from external knowledge sources and enhance BERT’s language understanding and reasoning capabilities. ExBERT takes full advantage of contextual word representations obtained from BERT and employs them to retrieve relevant external knowledge from knowledge graphs and to encode the retrieved external knowledge. Our model adaptively incorporates the external knowledge context required for reasoning over the inputs. Extensive experiments on the challenging SciTail and SNLI benchmarks demonstrate the effectiveness of ExBERT: in comparison to the previous state-of-the-art, we obtain an accuracy of $95.9\%$ on SciTail and $91.5\%$ on SNLI.

\end{abstract}

\section{Introduction}
Natural Language Inference (NLI), also known as Recognising Textual Entailment, is formulated as a - `directional relationship between pairs of text expressions, denoted by T (the entailing ``Text'') and H (the entailed “Hypothesis”). Text T, entails hypothesis H, if humans reading T would typically infer that H is most likely true.' \cite{rteChallenge}. 
\begingroup
\begin{table}[!h]
    \vspace{-4mm}
    \caption{SNLI premise (\textbf{P}) \& hypothesis (\textbf{H}) and commonsense triples (red) from ConceptNet KG. Commonsense knowledge enrich premise-hypothesis context and helps the NLI model in reasoning.}
    \renewcommand{\arraystretch}{1.1}
    \centering
    \begin{tabular}{p{0.95\columnwidth}} \\ \hline

    \textbf{P:} Four boys are about to be hit by an approaching \textcolor{red} {wave}. \textcolor{red}{(wave RelatedTo crash)} \\ \hline
     
     \textbf{H:} A giant wave is about to \textcolor{red}{crash} on some boys. \textcolor{red}{(crash IsA hit)} \\ \hline
     
    \end{tabular}

    \label{tab:snliExamples}
    \vspace{-0.6 cm}
\end{table}
\endgroup
The NLI task definition relies on common human understanding of language and real-world commonsense knowledge. NLI is a complex reasoning task, and NLI models can not rely solely on training data to acquire all the real-world commonsense knowledge required for reasoning and inference \cite{chen2018extknow}. For example, consider the premise-hypothesis pair in Table \ref{tab:snliExamples}, if the training data do not provide the common knowledge that, \textit{(wave RelatedTo crash)} and \textit{(crash IsA hit)}, it will be hard for the NLI model to correctly recognise that the premise entails the hypothesis.

Recently, deep pre-trained language representations models (PTLMs) such as BERT \cite{bert2019devlin} achieved impressive performance improvements on a wide range of NLP tasks. These models are trained on large amounts of raw text using a self-supervised language modelling objective. Although pre-trained language representations have significantly improved the state-of-the-art on many complex natural language understanding tasks, they lack grounding to real-world knowledge and are often unable to remember real-world facts when required \cite{needCSKnow2019Kwon}. Investigations into the learning capabilities of PTLMs reveal that the models fail to recall facts learned at training time, and do not generalise to rare/unseen entities \cite{baracks2019logan}. Knowledge probing tests \cite{needCSKnow2019Kwon} on the commonsense knowledge of ConceptNet \cite{conceptNet2017speer} reveal that PTLMs such as BERT have critical limitations when solving problems involving commonsense knowledge. Hence, infusing external real-world commonsense knowledge can enhance the language understanding capabilities of PTLMs and the performance on the complex reasoning tasks such as NLI.

Incorporating external commonsense knowledge into pre-trained NLI models is challenging. The main challenges are (i) \textbf{Structured Knowledge Retrieval:} Given a premise-hypothesis pair how to effectively retrieve specific and relevant external knowledge from the massive amounts of data in Knowledge Graphs (KGs). Existing models \cite{chen2018extknow}, use heuristics and word surface forms of premises-hypothesis which may be biased and the retrieved knowledge may not be contextually relevant for reasoning over premise-hypothesis pair. (ii) \textbf{Encoding Retrieved Knowledge:} Learning the representations of the retrieved external knowledge amenable to be fused with deep contextual representations of premise-hypothesis is challenging. Various KG embedding techniques \cite{kgEmbedSurvey2017Wang} are employed to learn these representations. However, while learning, the KG embeddings are required to be valid within the individual KG fact and hence might not be predictive enough for the downstream tasks \cite{kgEmbedSurvey2017Wang} (iii) \textbf{Feature Fusion:} How to fuse the learned external knowledge features with the premise-hypothesis contextual embeddings. This feature fusion requires substantial NLI model adaptations with marginal performance gains \cite{chen2018extknow}.

To overcome the aforementioned challenges, we propose, \textbf{ExBERT} - an External knowledge enhanced BERT model which enhances the contextual representations of BERT model with external commonsense knowledge to improve BERT's real-world grounding and reinforce its reasoning and inference capabilities. ExBERT utilises BERT for learning the contextual representation of premise-hypothesis as well as the representations of retrieved external knowledge. The aim here is to take full advantage of contextual word representations obtained from pre-trained language models and the real-world commonsense knowledge from Knowledge Graphs (KGs).

The main contributions of this paper are: (i) we devise a new approach, ExBERT, to incorporate external knowledge in contextual word representations. (ii)  we investigate and demonstrate the feasibility of using contextual word representation for encoding external knowledge obviating learning specialised KG embeddings. To the best of our knowledge, this is the first study of its kind, indicating a potential future research direction. (iii) we introduce a new external knowledge retrieval mechanism for NLI that is capable of retrieving fine-grained contextually relevant external knowledge from KGs.

\section{Related Work}
\label{sec:relatedWork}
\textbf{Traditional Attention-Based Models} do not utilise contextual representations from PTLMs \cite{gajbhiye2018CAM}. KG-Augmented Entailment System (KES) \cite{infusing2019kapanipathi} augments the NLI model with external knowledge encoded using graph convolutional networks. Knowledge-based Inference Model (KIM) \cite{chen2018extknow} incorporates  lexical-level knowledge (such as synonym and antonym) into its attention and composition components. ConSeqNet \cite{conseqNet2019wang}, a system of a text-based model and a graph-based model, concatenates the output of the two models, to be fed to a classifier. The {A}dv{E}ntu{R}e \cite{adventure2018khot} framework train the decomposable attention model with adversarial training examples generated by incorporating knowledge from linguistic resources, and with a sequence-to-sequence neural generator. BiCAM models improve the performance of NLI baselines via the incorporation of knowledge from the ConcepNet and Aristo Tuple KGs by factorised bilinear pooling \cite{gajbhiye2020BiCAM}.  

\textbf{PTLM-Based Models} PTLM-based models such as OpenAI GPT \cite{gpt2018radford} and BERT \cite{bert2019devlin} leverage transfer learning from a large textual corpus and are fine-tuned on NLI datasets. Specifically, OpenAI GPT pre-trains the Transformer \cite{attention2017vaswani} model in an unsupervised manner with the standard language modelling objective and fine-tunes the model in a supervised manner for NLI. Semantics-aware BERT (SemBERT) \cite{semanticsawareBERT2020zhang} demonstrated the benefit of enriching the BERT{'}s contextual representation with the semantic roles. BERT model has shown to be robust to adversarial examples when external knowledge is incorporated to the attention mechanism using simple transformations \cite{knowledgeEnhanced2019li}. The KES \cite{infusing2019kapanipathi} model highlighted above further evaluates their system with BERT contextual embeddings in the framework.

\section{Methodology}
\label{sec:methodology}
ExBERT architecture is depicted in Fig.\ref{fig:exBERT}. In this section, we describe the key components of ExBERT and their detailed implementation including the model architecture in Section \ref{sec:modelArch}. We start by describing the contextual representation based external knowledge retrieval procedure in Section \ref{sec:knowRetrieval}.

\begingroup
\begin{figure}[!t]
  \centering
    \includegraphics[width=0.74\columnwidth]{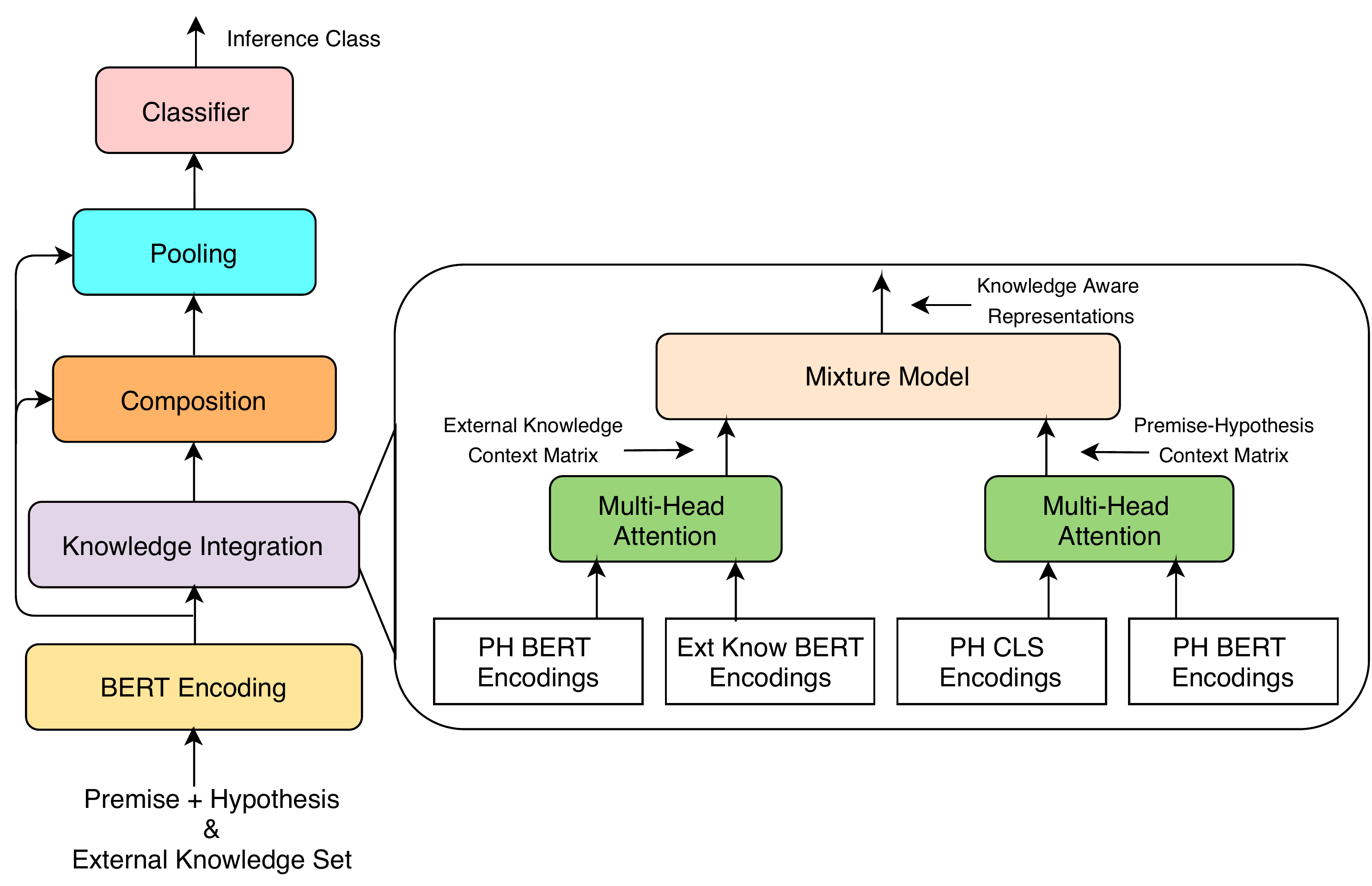}
    \caption{A high-level view of ExBERT architecture.}
    \label{fig:exBERT}
    \vspace{-0.5cm}
\end{figure}
\endgroup

\subsection{External Knowledge Retrieval: Selection and Ranking}
\label{sec:knowRetrieval}

Retrieval and preparation of contextually specific and relevant information from knowledge graphs are complex and challenging tasks. The crucial challenge is to identify the knowledge specific and relevant to the task at hand from the massive amounts of noisy KG data \cite{semanticSearch2016Bast}. Different from the previous approaches that use word surface forms to retrieve external knowledge, we use the cosine similarity between the contextual representations of premise-hypothesis tokens and the external knowledge. The external knowledge for the premise and hypothesis is retrieved individually. Below we explain the procedure for the premise. The same procedure is applied to the hypothesis. The output of external knowledge retrieval is the set of contextually relevant KG triples for the premise and hypothesis. We divide the external knowledge retrieval process into two parts: Selection and Ranking.  

\textbf{Selection} We first filter the stop words from the premise. Then we retrieve all the KB triples that contain the tokens of the premise as one of the words in the head entity of KG triples. For example, for the token ``\textit{speaking}'' one of the retrieved KG fact is ``\textit{public\_speaking IsA speaking}''. The retrieved triples are converted to a sentence. For example, the previous triple is transformed into ``\textit{public speaking is a speaking}''. The selection process retrieves a large number of KG triples which are not all relevant to the context of the premise. We filter the selected triples in the ranking step.

\textbf{Ranking} step ranks the selected KG triples according to the contextual similarity to the fine-grained context of the premise. Specifically, given the BERT generated context-aware representation of the premise tokens, we group all the bigrams of the representations. Each group of the bigram representation is averaged, and the cosine similarity is calculated with the average of the BERT representation of each of the selected KG triple sentence (created in selection step). For each bigram, we choose the KG triple sentence with the highest cosine similarity score.
To capture the fine-grained context of the premise, we repeat the ranking step with the trigrams, fourgrams, and the average of the whole premise BERT representations and retrieve the KG record with highest cosine similarity for each of the grams. Duplicate KG triple sentences are removed form the final set of retrieved knowledge. 

The advantages of our knowledge retrieval mechanism are that it is free from heuristic biases, requires no feature engineering, and the retrieved external knowledge is contextually relevant to the fine-grained context of the premise and hypothesis.

\subsection{Model Architecture}
\label{sec:modelArch}

\textbf{BERT Encoding Layer} This layer uses the BERT encoder to learn the context-aware representations of premise-hypothesis and the set of retrieved external knowledge. 

Specifically, given premise $P = \{p_i\}_{i=1}^n$, hypothesis $H = \{h_j\}_{j=1}^m$, and the set of external knowledge  $EXT = \{\{e_r\}_{r=1}^l\}^k$, where $r$ is the number of tokens in the external knowledge and $k$ is the number of retrieved external knowledge. For encoding the premise and hypothesis, we input $P$ and $H$ to $\mathrm{BERT}$ in the following form 
\begin{align}
\centering
    S & = [\langle \mathrm{CLS} \rangle, P, \langle \mathrm{SEP} \rangle,  H, \langle \mathrm{SEP} \rangle] \\
    H & = \mathrm{BERT} (S) \in \mathbb{R}^{(n+m+3) \times h}
\end{align}

where $\langle \mathrm{SEP} \rangle$ is the token separating $P$ and $H$, $\langle \mathrm{CLS} \rangle$ is the classification token, and $h$ is the dimension of the hidden state. When the BERT model is fine-tuned for the downstream task, the fine-tuned hidden state vector ($\mathbf{h}_{cls}$) corresponding to the classification token is used as the aggregate representation for the sequence. 
For each of the external knowledge in the set $EXT$, we generate the context-aware representations using the same $\mathrm{BERT}$ encoder as used for premise-hypothesis above as follows 
\begin{align}
\label{eq:ek}
    {EXT}^k & =  \noindent{[\langle \mathrm{CLS} \rangle, e_1, \ldots, e_l, \langle \mathrm{SEP} \rangle]} \\
    {E}^k &  = \mathrm{BERT}({EXT}^k) \in \mathbb{R}^{l+2 \times h} \\
    \mathbf{e}^k  &= \mathrm{Mean Pooling}({E}^k) \in \mathbb{R}^{1 \times h}
\end{align}

The averaged context-aware vector representation ($\mathbf{e}$) generated for each of the ($k$) retrieved external knowledge are stacked to create the context-aware matrix, ${E} \in \mathbb{R}^{k \times h}$.

\textbf{Knowledge Integration Layer} This layer integrates external knowledge into the premise-hypothesis contextual representations by means of multi-head dot product attention. The layer uses a mixture model to allow a better trade-off between the context from external knowledge and the premise-hypothesis context \cite{leveraging2017Yang}. The mixture model learns two parameter matrices, $A$ and $B$, weighing the importance of premise-hypothesis context and the external knowledge context.

\textbf{Multi-head Attentions} To measure the importance of  external knowledge to each context-aware premise-hypothesis representation, we apply multi-head dot product attention \cite{attention2017vaswani} between the context-aware representations of external knowledge and that of premise-hypothesis.

In multi-head dot product attention, the context-aware representations are projected linearly to generate the queries, keys and values. As we use the multi-head attention to highlight the external knowledge important to premise-hypothesis context, premise-hypothesis representation ($H$) generates the query matrix (${H}^q$) via linear projection and the two linear projections of external knowledge representation ($E$) generate the keys and values. The attention function is defined as
\begin{equation}
\label{eq:attExtKnowPH}
    \mathrm{Attention} (H^q, E^k, E^v) = \mathrm{softmax} (\frac{{H^q}{E^k}^T}{\sqrt{h_k}}){E}^v
\end{equation}
Then the multi-head attention is 
\begin{equation}
\begin{split}
\label{eq:multiHead}
    {C}_{ph}^{ext} &= \mathrm{MH}(H^q, E^k, E^v) \\ &= \mathrm{Concat(head_1, \ldots, head_h)} W^o
    \end{split}
\end{equation}

where $\mathrm{head_i} = \mathrm{Attention}(H^q{W}_{i}^{q}, E^k{W}_{i}^{k}, E^v{W}_{i}^{v})$ and ${W}_{i}^{q}, {W}_{i}^{k}, {W}_{i}^{v},$ and $W^o$ are projection matrices and $i$ is the number of attention heads ($12$ in our case). The output of multi-head attention, ${C}_{ph}^{ext} \in \mathbb{R}^{(m + n + 3) \times h}$ is an attention-weighted context matrix measuring the importance of the external knowledge context to each of the context-aware premise-hypothesis representation.

Similarly, to measure the importance of each premise-hypothesis BERT representation ($H$) to the aggregate premise-hypothesis representation (hidden representation $\mathbf{h}_{cls}$ corresponding to $\mathrm{CLS}$ token), we apply the multi-head attention between $\mathbf{h}_{cls}$ token representation and $H$ as 
\begin{equation}
    \mathrm{Attention} ({C}_{cls}^{q}, H^k, H^v) = \mathrm{softmax} (\frac{{C}_{cls}^{q}{H^k}^T}{\sqrt{h_k}}){H}^v
\end{equation}

where ${C}_{cls}^{q} \in \mathbb{R}^{(m + n + 3) \times h}$ is a matrix obtained by repeating $\mathbf{h}_{cls}$ hidden state $(n + m + 3)$ number of times. The multi-head attention is calculated similar to (Eq. \ref{eq:multiHead}) that outputs a context matrix ${C}_{ph}^{cls} \in \mathbb{R}^{(m + n + 3) \times h}$. The output matrix ${C}_{ph}^{cls}$ is an attention-weighted context matrix measuring the importance of each of the premise-hypothesis representation to the aggregate premise-hypothesis representation. 

\textbf{Mixture Model}
The mixture model learns a trade-off between the premise-hypothesis context and the context from external knowledge and is defined as
\begin{equation}
    M = A {C}_{ph}^{ext} + B {C}_{ph}^{cls}
\end{equation}
where $A$ and $B$ are the parameter metrices, learned with a single layer neural network and $ A + B = J \in \mathbb{R}^{(n +m +3) \times 1}$, $J$ is a matrix of all ones. The parameters $A$ and $B$ learn to balance the proportion of incorporating the premise-hypothesis context and the context from external knowledge. Each of the representations in $M \in \mathbb{R}^{(m + n + 3) \times h}$ can be regarded as a knowledge aware state representation that encodes external knowledge context information with respect to the context of each of the premise-hypothesis representation.

\textbf{Composition Layer} We compose the knowledge state representation ($M$) to the corresponding premise-hypothesis representation to obtain knowledge-aware matrix $\widehat{H}$ as
\begin{equation}
    \widehat{H} = H + M
\end{equation}

\textbf{Pooling Layer} The pooling layer creates a fixed-length representation from premise-hypothesis representations $H$ and the knowledge-aware representations $\widehat{H}$. We apply the standard mean and max pooling mechanisms as
\begin{equation}\label{eq:fixedPremise}
    \mathbf{h}_{mean}           =  \mathrm{MeanPooling} (H) \qquad
    \mathbf{h}_{max}            =  \mathrm{MaxPooling} (H) 
\end{equation}
\begin{equation}\label{eq:fixedPremise}
    \hat{\mathbf{h}}_{mean}     =  \mathrm{MeanPooling} (\widehat{H}) \qquad
    \hat{\mathbf{h}}_{max}      =  \mathrm{MaxPoolling} (\widehat{H}) 
\end{equation}
\textbf{Classification Layer} We classify the relationship between premise and hypothesis using a Multilayer Perceptron (MLP) classifier. The input to the MLP is the concatenation of pooled representations as
\begin{equation}
    \mathbf{f} = [\mathbf{h}_{mean} ; \hat{\mathbf{h}}_{mean} ; \mathbf{h}_{max} ; \hat{\mathbf{h}}_{max}]
\end{equation}
The MLP consists of two hidden layers with $\mathrm{tanh}$ activation and a $\mathrm{softmax}$ output layer to obtain the probability distribution for each class. The network is trained in an end-to-end manner using multi-class cross-entropy loss.

\section{Experiments}
\subsection{Datasets}
\textbf{NLI \& KGs} The key contribution of this paper is the unique method to incorporate external knowledge into the pre-trained BERT representations. ExBERT is capable of incorporating knowledge from any external knowledge source that allows the knowledge to be retrieved, given an entity as input. This includes KBs with \textit{(head, relation, tail)} graph structure, KBs that contain only entity metadata without a graph structure and those that combine both a graph and entity metadata. 

In this work, we retrieve external commonsense knowledge from ConceptNet \cite{conceptNet2017speer} for evaluating ExBERT on SNLI (570,000 examples) \cite{snliData} and SciTail (27,000 examples)\cite{scitail} benchmarks, and from the science domain-targeted KG, Aristo Tuple  \cite{aristoTuple2017Dalvi} for evaluation on science domain SciTail dataset. 

ConceptNet is a multilingual KG comprising of 83 languages. We pre-process the ConceptNet data to retrieve the facts with head and tail entities in the English language. The final pre-processed ConceptNet that we retrieve the external knowledge from contains 3,098,816 ($\approx 3\textrm{M}$) commonsense facts connected by $47$ relations. Aristo Tuple is an English language KG that contains $294,000$ science domain facts connected with $955$ unique relations. We search the whole Aristo Tuple KG to retrieve relevant external knowledge.      

\subsection{Experimental Setup}
Following our external knowledge retrieval mechanism discussed in Section \ref{sec:knowRetrieval}, we first retrieve the external knowledge from ConceptNet and Aristo Tuple for SNLI and SciTail datasets via selection and ranking steps. In the ranking step, the English uncased $\mathrm{BERT_{BASE}}$ \cite{bert2019devlin} model is employed in feature extraction mode (i.e. without fine-tuning) to learn the contextual representations of the premise, the hypothesis and to each of the selected KG triple sentences. We then use the retrieved external knowledge to train the following three versions of ExBERT.

\textbf{Models} We used the English uncased $\mathrm{BERT_{BASE}}$  to train three versions of ExBERT: Two ExBERT+ConceptNet models on SNLI and SciTail respectively and one ExBERT+AristoTuple model on SciTail. The models utilise the external knowledge from the KG their name is suffixed. 



\textbf{Training Details} ExBERT is implemented in PyTorch using the base implementation of BERT\footnote{\url{https://github.com/huggingface/transformers}}. The underlying BERT is initialised with the pre-trained BERT parameters and follows the same fine-tuning procedure as the original BERT. During training, the pre-trained BERT parameters are fine-tuned with the other ExBERT parameters. We use the Adam optimiser \cite{adam2015Kingma} with the initial learning rate fine-tuned from $\{8e$-$6$, $2e$-$5$, $3e$-$5$, $5e$-$5\}$ and warm-up rate of $0.1$. The batch size is selected from $\{16, 24, 32\}$. The maximum number of epochs is chosen from $\{2, 3, 4, 5\}$. Dropout ratio of $0.5$ is used at the classification layer \cite{dropoutnli2018gajbhiye}. Texts are tokenised using word pieces, with a maximum length of $40$ for SNLI, $60$ for SciTail, and $15$ for external knowledge. The hyper-parameters are fine-tuned on the dev set of each NLI dataset.

\section{Results}
\label{sec:results}
The results of top-performing models on the SNLI\footnote{\url{https://nlp.stanford.edu/projects/snli/}} and SciTail\footnote{\url{https://leaderboard.allenai.org/scitail/submissions/public}} leaderboards are summarised in Table \ref{tab:mainResults}. For fairness of comparison, we compare ExBERT with only the PTLMs based NLI models that leverage external knowledge. On \textbf{SNLI}, the performance of the state-of-the-art models is highly competitive.
\begin{table}
\caption{Results on SNLI and SciTail dataset: For SNLI, ExBERT uses ConceptNet KG. For SciTail ExBERT uses ConceptNet KG and AristoTuple KGs.} \label{tab:mainResults}
\vspace{+2mm}
\parbox{.50\linewidth}{
\centering

    \resizebox{0.5\columnwidth}!{
    \begin{tabular}{lc} \hline

        \multicolumn{2}{c}{\textbf{SNLI Dataset}} \\ \hline
        \multicolumn{2}{c}{\textbf{$\mathbf{BERT_{BASE}}$ as Base Model}} \\ \hline
        \textbf{NLI Model} & \textbf{Test Acc(\%)} \\ \hline 
        
        $\mathrm{BERT}$
         $\mathrm{BERT_{BASE}}$ + SRL \cite{explicitContext2019zhang}   &   89.6          \\
         OpenAI GPT \cite{gpt2018radford}                               &   89.9          \\
         $\mathrm{BERT_{BASE}}$ \cite{knowledgeEnhanced2019li}          &   90.5          \\
         $\mathrm{BERT_{BASE}}$ \cite{multitaskDNN2019liu}              &   90.8          \\
         BERT+LF \cite{improvePretrainedParser2019Deric}                &   90.5          \\
         $\mathrm{SemBERT_{BASE}}$ \cite{semanticsawareBERT2020zhang}   &   91.0          \\
         $\mathrm{MT-DNN_{BASE}}$ \cite{multitaskDNN2019liu}            &   91.1          \\
         MT-DNN+LF \cite{improvePretrainedParser2019Deric}              &   91.1          \\ \hline
         
         \multicolumn{2}{c}{\textbf{$\mathbf{BERT_{LARGE}}$ as Base Model}} \\ \hline
         
         $\mathrm{BERT_{LARGE}}$ \cite{multitaskDNN2019liu}             &   91.0          \\
         $\mathrm{BERT_{LARGE}}$ + SRL \cite{explicitContext2019zhang}  &   91.3          \\
         $\mathrm{SemBERT_{LARGE}}$ \cite{semanticsawareBERT2020zhang}  &   91.6          \\
         $\mathrm{MT-DNN_{LARGE}}$ \cite{multitaskDNN2019liu}           &   91.6          \\ \hline
         \textbf{ExBERT+ConceptNet (Ours)}                              & \textbf{91.5}   \\ \hline

    \end{tabular}
    }
}%
\hfill
\parbox{.50\linewidth}{
\centering
    \resizebox{0.50\columnwidth}!{
    \begin{tabular}{lc} \hline

        \multicolumn{2}{c}{\textbf{SciTail Dataset}} \\ \hline
        \multicolumn{2}{c}{\textbf{$\mathbf{BERT_{BASE}}$ as Base Model}} \\ \hline
        
        \textbf{NLI Model} & \textbf{Test Acc\%)} \\ \hline
        
        OpenAI GPT \cite{gpt2018radford}                        & 88.3           \\
        $\mathrm{BERT_{BASE}}$ \cite{multitaskDNN2019liu}       & 92.5           \\
        BERT+LF \cite{improvePretrainedParser2019Deric}         & 92.8           \\
        $\mathrm{MT-DNN_{BASE}}$ \cite{multitaskDNN2019liu}     & 94.1           \\
        MT-DNN+LF\cite{improvePretrainedParser2019Deric}        & 94.3           \\  \hline

         \multicolumn{2}{c}{\textbf{$\mathbf{BERT_{LARGE}}$ as Base Model}} \\ \hline
         
         $\mathrm{BERT_{LARGE}}$ \cite{multitaskDNN2019liu}     & 94.4           \\
         $\mathrm{MT-DNN_{LARGE}}$ \cite{multitaskDNN2019liu}   & 95.0           \\  \hline
         
         \textbf{ExBERT+ConceptNet (Ours)}                      &  \textbf{95.2} \\
         \textbf{ExBERT+AristoTuple (Ours)}                     &  \textbf{95.9} \\ \hline
         
    \end{tabular}
    }
}
\vspace{-6mm}
\end{table}

We observe that ExBERT outperforms all the existing baselines on the SNLI dataset and pushing the benchmark to $91.5\%$ within the models using $\mathrm{BERT_{BASE}}$ as the base model. ExBERT achieves a maximum performance improvement of $1.9\%$ over the previous state-of-the-art $\mathrm{BERT_{BASE}}$ + SRL \cite{explicitContext2019zhang} baseline.

Among the models built on $\mathrm{BERT_{LARGE}}$ with more than 340M million parameters \cite{bert2019devlin}, our ExBERT\footnote{We expect further improvements in ExBERT{'}s performance with $\mathrm{BERT_{LARGE}}$, however we left the evaluation for future work due to the limited computing resources.} with $\mathrm{BERT_{BASE}}$ (110M parameter) remarkably outperforms the $\mathrm{BERT_{LARGE}}$ and $\mathrm{BERT_{LARGE}}$ + SRL \cite{explicitContext2019zhang} models with the absolute improvements of $0.5\%$ and $0.2\%$ respectively, and is able to match the performance of $\mathrm{SemBERT_{LARGE}}$ \cite{semanticsawareBERT2020zhang} and $\mathrm{MT-DNN_{LARGE}}$ \cite{multitaskDNN2019liu} models. 

On \textbf{SciTail} (Table \ref{tab:mainResults}), ExBERT outperforms all the existing models including the models built on $\mathrm{BERT_{LARGE}}$ model. Our best performing model, ExBERT+AristoTuple demonstrates an absolute improvement of $7.6\%$ over the established baseline of OpenAI GPT \cite{gpt2018radford}. Moreover, using only $\mathrm{{BERT}_{BASE}}$ as the underlying model, our ExBERT+AristoTuple outperforms $\mathrm{BERT_{LARGE}}$ based $\mathrm{MT-DNN_{LARGE}}$ \cite{multitaskDNN2019liu} model by $0.9\%$. 

We observe higher performance improvements on the smaller SciTail dataset which demonstrates that incorporating external knowledge helps the model with small training data. Further, we observe that ExBERT attains higher accuracy when external knowledge is incorporated from the science domain-specific KG, Aristo Tuple as compared to when external knowledge is added from the commonsense KG, ConceptNet. The specialised scientific knowledge in Aristo Tuple is more beneficial to SciTail.
\vspace{-0.2 cm}
\section{Analysis}
\label{sec:analysis}
\vspace{-0.2 cm}
\subsection{Number of External Features}
\label{sec:extFeature}

\begin{figure}[h]
\vspace{-0.45cm}
    \centering
    \includegraphics[height=4cm]{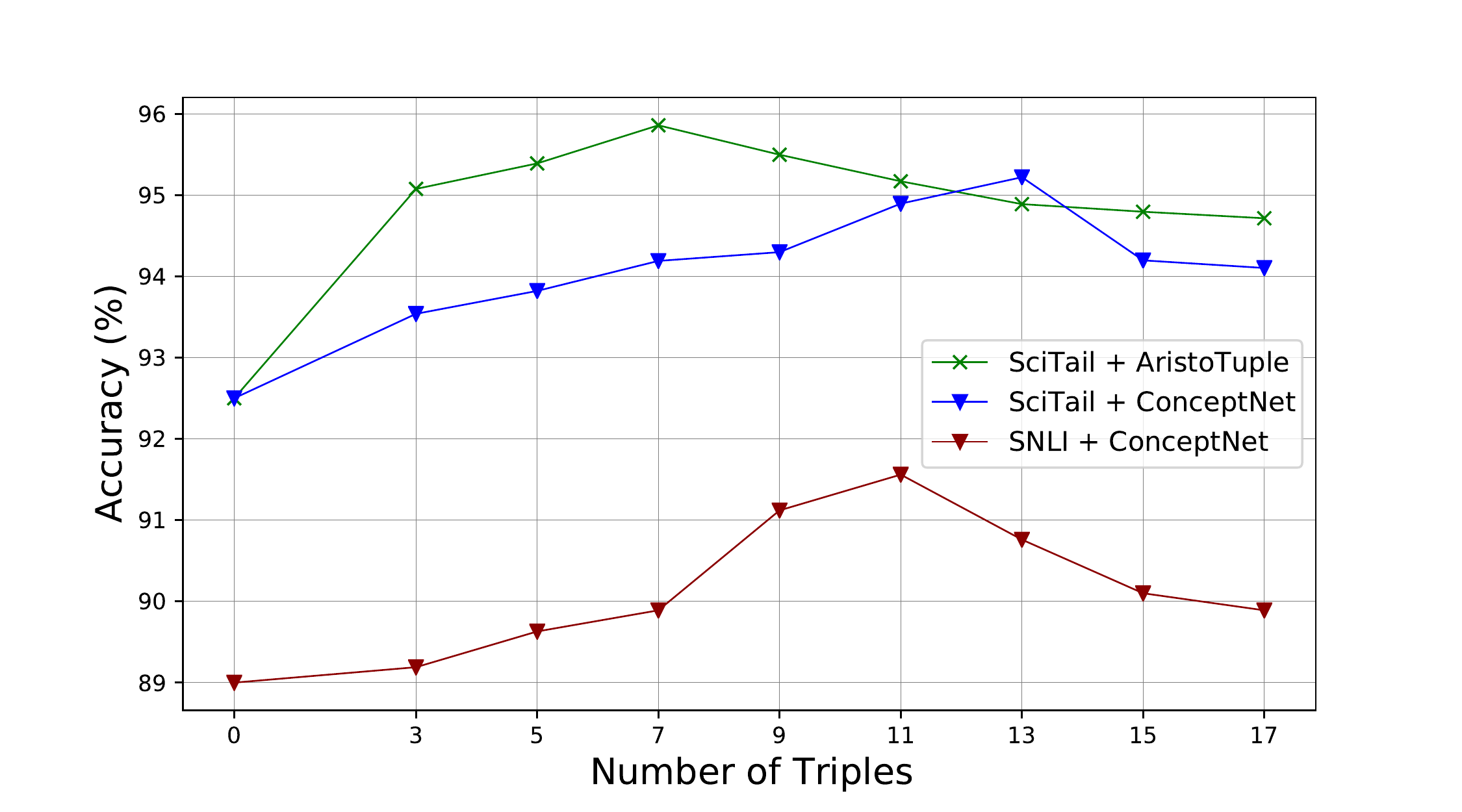}
    \caption{ExBERT accuracy with varying amount of external knowledge.}
    \label{fig:lenPlots}
    \vspace{-0.5cm}
\end{figure}

To investigate the effect of incorporating various numbers of external knowledge features, we vary the number of KG triple sentences input to ExBERT. Particularly, we are interested in answering the question: How many commonsense features are required for the optimal model performance? Figure \ref{fig:lenPlots} illustrates the results of the experiment. \textbf{For SNLI}, ExBERT achieves the highest accuracy $(91.5\%)$ using $11$ external knowledge sentences. We observe a decrease in accuracy when increasing the number of external knowledge sentences after $11$. The fewer number of external knowledge sentences required, compared to SciTail dataset, to achieve the maximum accuracy on SNLI dataset, is attributed to the limited linguistic and semantic variation and the short average length of stop-word filtered premise ($7.35$ for entailment and neutral class) and hypothesis ($3.61$ for entailment and $4.45$ for neutral class) \cite{scitail} of the SNLI dataset, which limits its ability to fully extract and exploit external KG knowledge.

\textbf{For SciTail}, ExBERT when evaluated using the general commonsense knowledge source ConceptNet, requires a relatively high number of external knowledge sentences $(15)$ to achieve the maximum accuracy. This is due to the higher syntactic and semantic complexity of SciTail, which needs more knowledge to reason. However, when evaluated with the domain-specific Aristo Tuple KG, the model achieves the highest accuracy with fewer $(7)$ external knowledge sentences. To reiterate, domain specific knowledge in Aristo Tuple improves the model performance with less external knowledge.

\subsection{Qualitative Analysis}
\subsubsection{Case Study}
This section provides the case study of different premise-hypothesis pairs and the corresponding external knowledge, to vividly show the effectiveness of ExBERT in adaptively identifying the relevant features from the supplied external knowledge. Recall that given a context-aware representation of premise-hypothesis token, the relevance of the retrieved external knowledge in $E$ is measured by the multi-head attention defined in Eq.(\ref{eq:attExtKnowPH}). We average the attention weights of all heads and plot a heat map. 

\begin{figure*}[!h]
\vspace{-0.45cm}
    \centering
    \includegraphics[width=0.6\columnwidth]{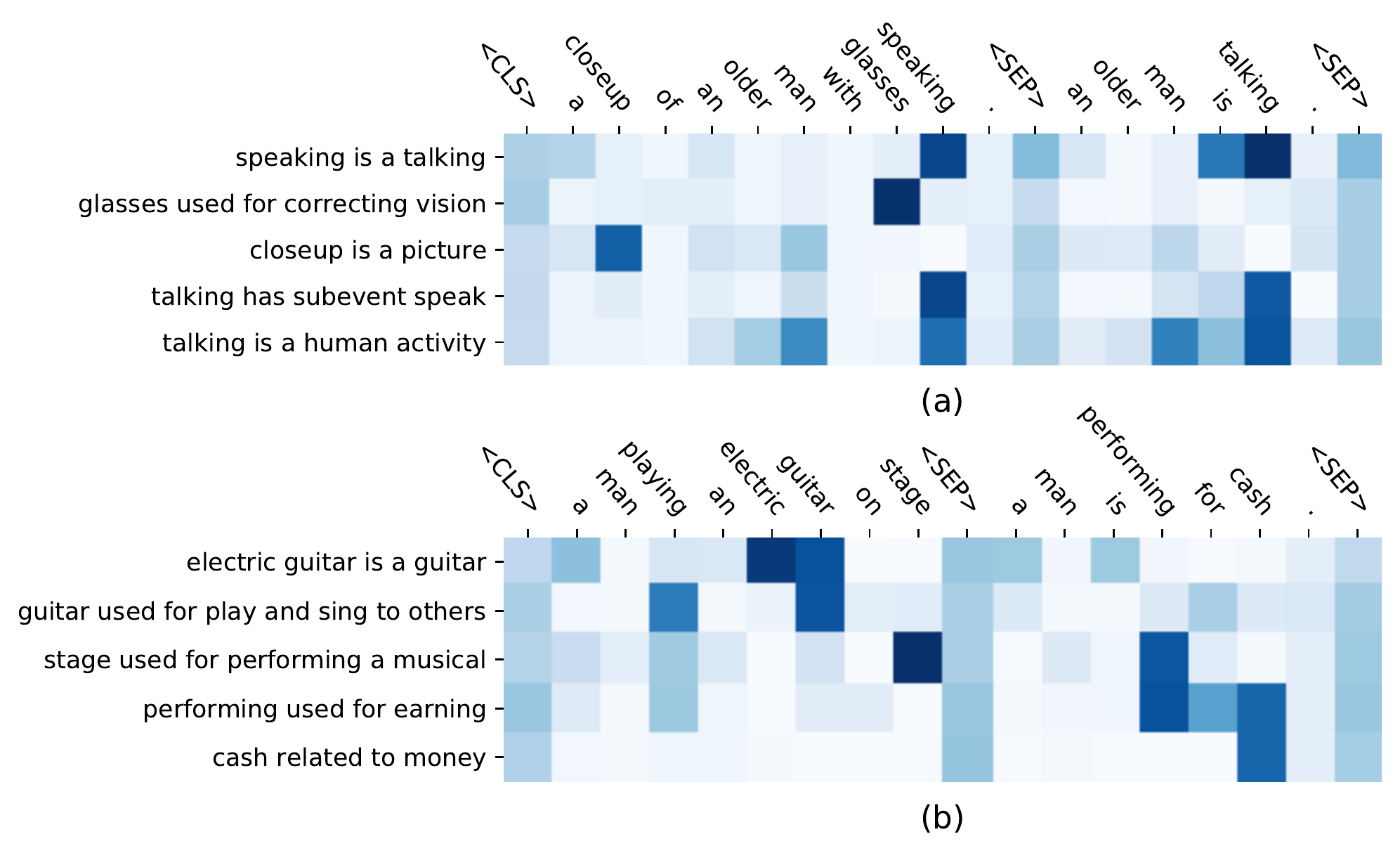}
    \caption{ Case Study. Visualisation of ExBERT's attention between external knowledge from ConceptNet (y axis) and SNLI premise-hypothesis pair tokens (x axis).}
    \label{fig:attenPlots}
    \vspace{-0.5cm}
\end{figure*}

Fig.\ref{fig:attenPlots} presents the heat map showing the attention of premise-hypothesis tokens to the retrieved external knowledge sentences from ConceptNet. In Fig.\ref{fig:attenPlots}(a), we can see, these attention distribution is quite meaningful, with the ``\textit{speaking}'' and ``\textit{talking}'' attending mainly to the retrieved  external knowledge ``\textit{speaking is talking}''. Similarly, the tokens ``\textit{speaking}'' ``\textit{talking}'' and ``\textit{man}'' attends to ``\textit{talking is a human activity}''. In Fig.\ref{fig:attenPlots}(b) among the other attentions, the most prominent can be observed between the tokens ``\textit{performing for cash}'' and the external knowledge sentence ``\textit{performing used for earning}''.

Attending to the relevant external knowledge demonstrates the ExBERT's ability to effectively utilise the retrieved external knowledge based on the context from the premise and hypothesis.

\section{Conclusion}
We introduced ExBERT to enrich the contextual representation of BERT with real-world commonsense knowledge from external knowledge sources and to enhance its language understanding and reasoning capabilities. 
ExBERT can incorporate external knowledge from any external knowledge source that allows the knowledge to be retrieved, given an entity. We devised a novel external knowledge retrieval mechanism utilising contextual representations to retrieve relevant external knowledge. Experimental results on SNLI and SciTail NLI benchmarks in conjunction with two KGs, ConceptNet and Aristo Tuple, shows that ExBERT achieves significant performance improvements over the previous state-of-the-art methods, including those which are enhanced by $\mathrm{{BERT}_{LARGE}}$. Further, we demonstrated the feasibility of utilising contextual representations for encoding external knowledge from KGs, which indicates a potential direction for future research.
\vspace{-0.5mm}
%
%
%
\bibliographystyle{splncs04}
\bibliography{emnlp2020}
\end{document}